\documentclass[letterpaper, 10 pt, journal, twoside]{ieeetran}

\IEEEoverridecommandlockouts                              

\usepackage{graphicx}
\usepackage{subcaption}
\usepackage[nolist]{acronym}
\usepackage{tikz}
\usepackage{pdflscape}
\usepackage{multirow}
\usepackage{siunitx}
\usetikzlibrary{fadings}
\usetikzlibrary{arrows.meta}
\usetikzlibrary{calc, arrows, decorations.markings, angles, quotes}
\usetikzlibrary{decorations.pathmorphing} 
\usetikzlibrary {decorations.fractals,spy}
\usepackage{xcolor}
\usepackage{xcolor,colortbl}
\usepackage{fixmath}
\usepackage[hidelinks]{hyperref}
\usepackage{amsmath}
\usepackage{pifont}

\usepackage{wasysym}
\newcommand{\yes}{\ding{51}}
\newcommand{\no}{\ding{55}}
\newcommand{\part}{\ding{108}}  
\newcommand{\partmark}{\LEFTcircle}  

\usepackage{tikz}
\usepackage{xcolor}
\usepackage{stfloats}

\usetikzlibrary{angles, quotes, calc, arrows.meta, patterns, decorations.markings, 3d}

\newcommand{\green}[1]{#1}

\author{Tim Lakemann$^{1}$, Daniel Bonilla Licea$^{2}$, Viktor Walter$^{1}$, Martin Saska$^{1}$
\thanks{Manuscript received: Aug. 05, 2026; Accepted Oct. 06, 2026.}
\thanks{This paper was recommended for publication by Editor Soon-Jo Chung upon evaluation of the Associate Editor and Reviewers' comments.
This work was funded by the CTU grant no SGS26/077/OHK3/1T/13, the Czech Science Foundation (GAČR) under research project no. 26-22419S, and  by the European Union under the project Robotics and advanced industrial production (reg. no. CZ.02.01.01/00/22\_008/0004590).}
\thanks{$^{1}$
Tim Lakemann, Viktor Walter and Martin Saska are with the Department of Cybernetics, Czech Technical University in Prague, 14900 Prague,
Czech Republic (e-mail: \protect\href{mailto:lakemtim@fel.cvut.cz}{lakemtim@fel.cvut.cz}).}%
\thanks{$^{2} $Daniel Bonilla Licea is with the Mohammed VI Polytechnic University in Morocco.}%
\thanks{Digital Object Identifier (DOI): see top of this page.}
}

\usepackage{amsmath,amsfonts,amssymb}

\title{Reflection-Based Relative Localization for Cooperative UAV Teams Using Active Markers}

\definecolor{ptmblue}{RGB}{000, 055, 125}
\definecolor{ptmlightblue}{RGB}{0, 136, 221}
\definecolor{ptmpink}{RGB}{255, 115, 182}
\definecolor{ptmpurple}{RGB}{102, 000, 255}
\definecolor{ptmgreen}{RGB}{164, 203, 141}
\definecolor{ptmorange}{RGB}{255, 157, 058}
\definecolor{ptmyellow}{RGB}{255, 232, 136}
\definecolor{ptmred}{RGB}{213, 048, 052}

\usepackage[normalem]{ulem}

\tikzset{
  invisible/.style={opacity=0},
  visible on/.style={alt={#1{}{invisible}}},
  alt/.code args={<#1>#2#3}{%
    \alt<#1>{\pgfkeysalso{#2}}{\pgfkeysalso{#3}} 
  },
}

\tikzset{
  camera/.pic = {
    \filldraw [black, fill=ptmyellow] (0.0,0.0) 
    -- (-0.6,1.3) 
    -- (0.6,1.3) 
    -- cycle;
    \filldraw [black,fill=ptmyellow] (-0.5,-0.5) rectangle (0.5,0.5);

  }
}

\tikzset{
  uav_frame/.pic = {
    \draw [line width=0.4mm] (-2,-1) -- (2,1);
    \draw [line width=0.4mm] (-2,1) -- (2,-1);
    \draw[] (-1.5,1.15) arc[start angle=20, end angle=270, x radius=0.5cm, y radius=0.4cm];
    \draw[] (1.5,1.15) arc[start angle=160, end angle=-90, x radius=0.5cm, y radius=0.4cm];
    \draw[] (-1.95,-0.6) arc[start angle=90, end angle=340, x radius=0.5cm, y radius=0.4cm];
    \draw[] (1.95,-0.6) arc[start angle=90, end angle=-160, x radius=0.5cm, y radius=0.4cm];

  }
}

\tikzset{
    rx/.pic = {
        \pic [rotate=0, scale=1] at (0, 0) {uav_frame};
        \pic [rotate=90, scale=1] at (0,0.5) {camera};
  }
}

\tikzset{
    tx/.pic = {
        \pic [rotate=0, scale=1] at (0, 0) {uav_frame};
        \filldraw[ptmred] (2,-1) circle (2mm);
        \filldraw[ptmred] (2,1) circle (2mm);
        \filldraw[ptmred] (-2,-1) circle (2mm);
        \filldraw[ptmred] (-2,1) circle (2mm);
    \filldraw (0,0) circle (5mm);

  }
}
\tikzset{
  uav_frame_flat/.pic = {
    \draw [line width=0.4mm] (-1.5,-1.5) -- (1.5,1.5);
    \draw [line width=0.4mm] (-1.5,1.5) -- (1.5,-1.5);
    \draw[] (-1.1,1.6) arc[start angle=0, end angle=270, x radius=0.5cm, y radius=0.5cm];
    \draw[] (1.1,1.6) arc[start angle=180, end angle=-90, x radius=0.5cm, y radius=0.5cm];
    \draw[] (-1.6,-1.1) arc[start angle=90, end angle=360, x radius=0.5cm, y radius=0.5cm];
    \draw[] (1.6,-1.1) arc[start angle=90, end angle=-180, x radius=0.5cm, y radius=0.5cm];

  }
}

\tikzset{
    rx_flat/.pic = {
        \pic [rotate=0, scale=1] at (0, 0) {uav_frame_flat};
        \pic [rotate=90, scale=1] at (0,0.) {camera};
  }
}
\newcommand{\PREPRINTYEAR}{2026}
\newcommand{\PUBLISHEDIN}{IEEE Robotics and Automation Letters}
\newcommand{\DOI}{10.1109/LRA.2026.3748684} 

\usepackage[placement=top,vshift=-7]{background}
\SetBgScale{1.0}
\SetBgContents{\PUBLISHEDIN. PREPRINT VERSION - DO NOT DISTRIBUTE.
\href{https://doi.org/\DOI}{DOI \DOI}}
\SetBgColor{black}
\SetBgAngle{0}
\SetBgOpacity{1.0}

\begin{document}

\onecolumn
{
  \topskip0pt
  \vspace*{\fill}
  \centering
  \LARGE{%
    \copyright{} \PREPRINTYEAR~\PUBLISHEDIN\\\vspace{1cm}
    Personal use of this material is permitted.
    Permission from \PUBLISHEDIN~must be obtained for all other uses, in any current or future media, including reprinting or republishing this material for advertising or promotional purposes, creating new collective works, for resale or redistribution to servers or lists, or reuse of any copyrighted component of this work in other works.}
    \vspace*{\fill}
}
\NoBgThispage
\twocolumn          	
\BgThispage

\bstctlcite{BSTcontrol}

\maketitle

\begin{abstract}
  Reflections of active markers in the environment are a common source of ambiguity in onboard visual relative localization.
This work presents a novel approach that exploits these typically unwanted reflections for onboard relative localization in heterogeneous multi-\ac{UAV} teams operating over specular, near-planar surfaces such as calm water and flat indoor floors.
Using only a camera, active markers, and a single-beam rangefinder, the method requires no prior knowledge of robot size, marker configuration, or surface reflectance properties, and propagates the uncertainty introduced by surface irregularities, including those of dynamic water surfaces.
We further characterize, analytically and experimentally, how surface tilt biases the estimate, and identify the slope range over which the level-surface assumption holds.
Indoor multi-\ac{UAV} and outdoor experiments validate the localization accuracy, demonstrating reliable operation across varying lighting conditions and greater effective range than state-of-the-art vision-based \green{active-marker} methods, with higher accuracy beyond 30 m.
The project website with the video is available at: \url{https://timlakemann.github.io/reflection-based-relative-localization/}.
\end{abstract}


\section{Introduction}
\IEEEPARstart{R}{eliable} relative localization among team members is crucial for multi-robot systems to collaboratively execute tasks safely and reliably \cite{Nguyen2023}.
In outdoor environments, \ac{GNSS} often lacks the reliability needed for close-formation flight.
Its signal accuracy and precision are particularly degraded in dense forests, natural landscapes, or urban canyons, and in some cases, it becomes entirely unavailable~\cite{Nguyen2023,zhangBestIntegerEquivariant2023}.
External reference systems such as \ac{RTK} antennas improve accuracy but constrain the operational area to the range of the ground station.

Onboard relative localization removes this dependence on external infrastructure.
Decentralized \ac{UWB} ranging provides centimeter-level distance accuracy~\cite{xunCREPESCooperativeRElative2023a}, but requires active radio exchange between team members, making it dependent on communication and susceptible to \ac{RF} jamming.
3D \ac{LIDAR} provides direct depth measurements robust to illumination, but it is heavy, power-intensive, and cannot associate a return with a specific team member~\cite{SwarmLIO2}.
Vision-based methods, in contrast, provide a cost-effective, compact, and lightweight solution that requires no communication between team members~\cite{xuDecentralizedVisualInertialUWBFusion2020a,schillingVisionBasedDroneFlocking2021}.
While passive markers like AprilTags improve localization accuracy, they impose constraints on system dynamics, which is particularly critical for \acp{UAV}~\cite{5979561}.
\begin{figure}[t]
    \centering
   {\scalebox{0.78}{\input{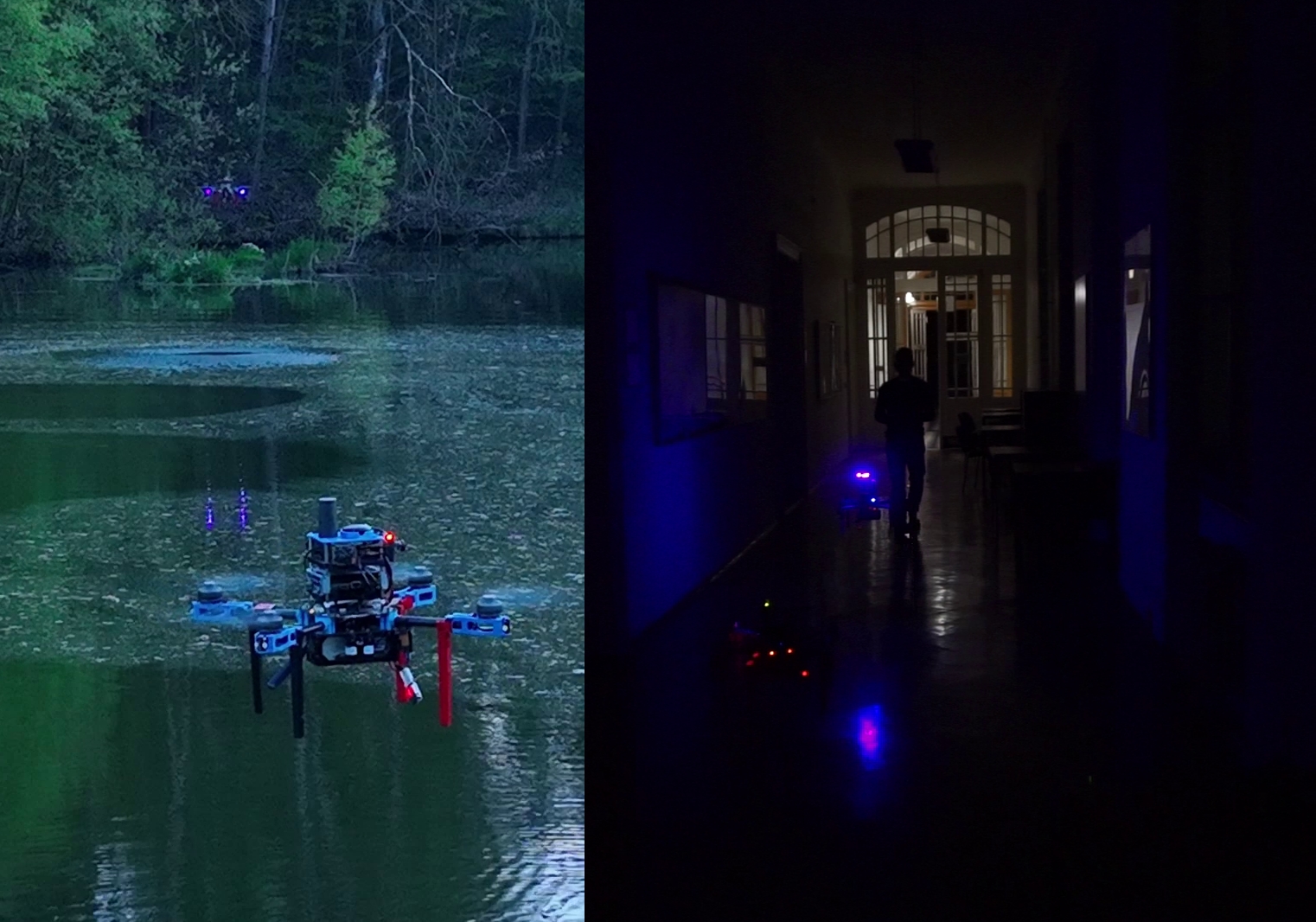}}}
   \caption{Outdoor (left) and indoor dark (right) experiments: A \ac{UAV} estimates the relative position by using surface reflections from active markers attached to a team member \ac{UAV}. Blue boxes: the \ac{UAV} with active markers attached. Red boxes: Surface reflections of the light emitted by the \ac{UAV}.}
   \label{fig:intro_img}
\end{figure}
Active markers overcome limitations related to marker size, enable reliable detection under varying lighting conditions, and, through unique blinking sequences, link each detection to a known robot identity~\cite{stuckeyRealTimeOpticalLocalization2024,walterUVDARSystemVisual2019}.
Such systems, however, infer relative distance from the pixel separation between markers, and therefore require prior knowledge of the marker configuration on each~\ac{UAV}~\cite{walterUVDARSystemVisual2019,faesslerMonocularPoseEstimation2014}.
As with marker-less or passive marker systems, the precision of this distance estimation decreases for \acp{MAV} with closely spaced markers.
The limited physical spacing reduces pixel separation as distance increases, eventually falling below the resolution threshold required for reliable distance estimation~\cite{walterUVDARSystemVisual2019}.

    This work addresses these limitations by introducing a reflection-based relative localization system for \ac{UAV} teams operating above specular, near-planar surfaces, such as calm water in marine environments or polished floors in indoor facilities (Fig.~\ref{fig:intro_img}).

Prior work has focused on mitigating surface reflections using polarization filters~\cite{bergerDepthStereoPolarization2017, yuanzhenliMultibaselineStereoPresence2002} or structured light~\cite{polarization_structured_light}.
Reflection-based localization has also been explored in the \ac{RF} spectrum~\cite{liMimoLocIndoorLocalization2024, liTriLocAccurateIndoor2023, zhangMultiPersonPassiveWiFi2023} and acoustic domains~\cite{anDiffractionReflectionAwareMultiple2022, birnieReflectionAssistedSound2020}.
In~\cite{7024911}, depth images are generated from water surface reflections.
However, the approach requires a stationary camera and 40 seconds per frame, making it impractical for collaborative \acp{UAV}.

To the best of our knowledge, relative localization through surface reflections remains unexplored for multi-UAV teams in the literature.
Unlike most vision-based relative localization methods, our approach exploits surface reflections to extend the operational range without requiring prior knowledge of the optical properties of the surface or of the size and marker configuration of the transmitting \ac{UAV}.

{By accounting for the spread of the reflection pattern caused by surface roughness (Fig.~\ref{fig:high_exposure}) and for sensor uncertainties, the approach remains reliable on surfaces that are not perfectly specular.
It also removes the marker-spacing constraint discussed above, allowing the compact marker configurations typically required on \acp{MAV}.}
Our contribution in this work is threefold:
\begin{enumerate}
        \item We propose a reflection-based relative localization approach that requires only a camera, active markers, and a single-beam rangefinder, with no prior knowledge of team member sizes, marker configurations, or optical surface properties.
        \item \green{We derive and experimentally validate the position bias induced by surface tilt, showing that it is asymmetric and that, for surfaces rising toward the transmitter, the estimates diverge at a finite range.}
          \item We validate the approach indoors and outdoors across heterogeneous platforms, demonstrating \green{onboard operation}, reliable performance under varying lighting, and improved accuracy over the state-of-the-art at ranges beyond \SI{30}{\meter}.
\end{enumerate}

\section{Reflection-Based Localization Method}\label{sec:method}
The proposed reflection-based relative localization method works effectively both indoors and outdoors, {regardless of \ac{UAV} size or prior knowledge of optical surface properties.}
Each \ac{UAV} emits a predefined unique blinking sequence via its \acp{LED}, generated following~\cite{liceaOpticalCommunicationbasedIdentification2025}, where the sequence design, Hamming-distance separation, and resulting crosstalk and association-error behavior are analyzed in detail.
The sequences are tracked and uniquely identified using the approach of~\cite{lakemann2025}, which was originally designed for low-exposure images in which each \ac{LED} appears as a small, bright dot of only a few pixels.
At higher exposure settings, required for reliable reflection detection, a single \ac{LED} may be segmented into multiple detected regions.
To address this, detections within a radius of $r_c$ pixels are grouped via a clustering step, with the dominant cluster, i.e., the one with the highest number of extracted signals, treated as the direct \ac{LED} source.
If an additional cluster exists and is located vertically below the dominant cluster, it is selected as the reflection candidate.
The position of each cluster is represented by its centroid, computed as the mean of all detections within the cluster.
In the following, $p_d$ denotes the centroid of the direct \ac{UAV}-emitted light source and $p_r$ the centroid of its reflection.
When the blinking sequence cannot be extracted from the reflected region, $p_r$ is instead identified geometrically as the detection located vertically below $p_d$ in the image.
Since the detected centroids are transformed into the gravity-aligned frame prior to any geometric reasoning, the vertical alignment criterion corresponds to physical verticality rather than image-axis alignment.
\begin{figure}[t]
   \centering
   {\scalebox{0.6}{\input{high_exposure_image}}}
   \caption{Excerpt from an outdoor experiment showing active markers attached to the \ac{UAV} (blue) and their diffuse reflections on the water surface (red), captured onboard the \ac{UAV}.}
   \label{fig:high_exposure}
        \vspace{-11.5pt}
\end{figure}
\begin{figure*}[ht!]
        \centering
        \subfloat[\label{fig:method:tikz_v_view}]{%
            \begin{tikzpicture}[scale=0.3] 
                \input{system_vertical_angles.tex} 
            \end{tikzpicture}
        }
        \hspace{1.5cm}
        \subfloat[\label{fig:method:tikz_h_view}]{%
           \begin{tikzpicture}[scale=0.7] 
               \input{tikz_horizontal_view.tex}
            \end{tikzpicture}
        }
        \caption{One \ac{UAV} equipped with a camera (yellow) constructs an elliptical cone based on the diffuse reflections of light emitted by the transmitting \ac{UAV} (top left).
        (a) Side view of our approach, showing the potential location of the transmitting \ac{UAV} determined by the intersection of two elliptical cones (gray area).
        (b) Top-down view of the observer \ac{UAV} with the ellipse around the diffuse reflection.}
        \vspace{-5pt}
        \label{fig:method:tikz_overview}
\end{figure*}

\begin{figure}[t]
   \centering
   \includegraphics[width=0.75\columnwidth]{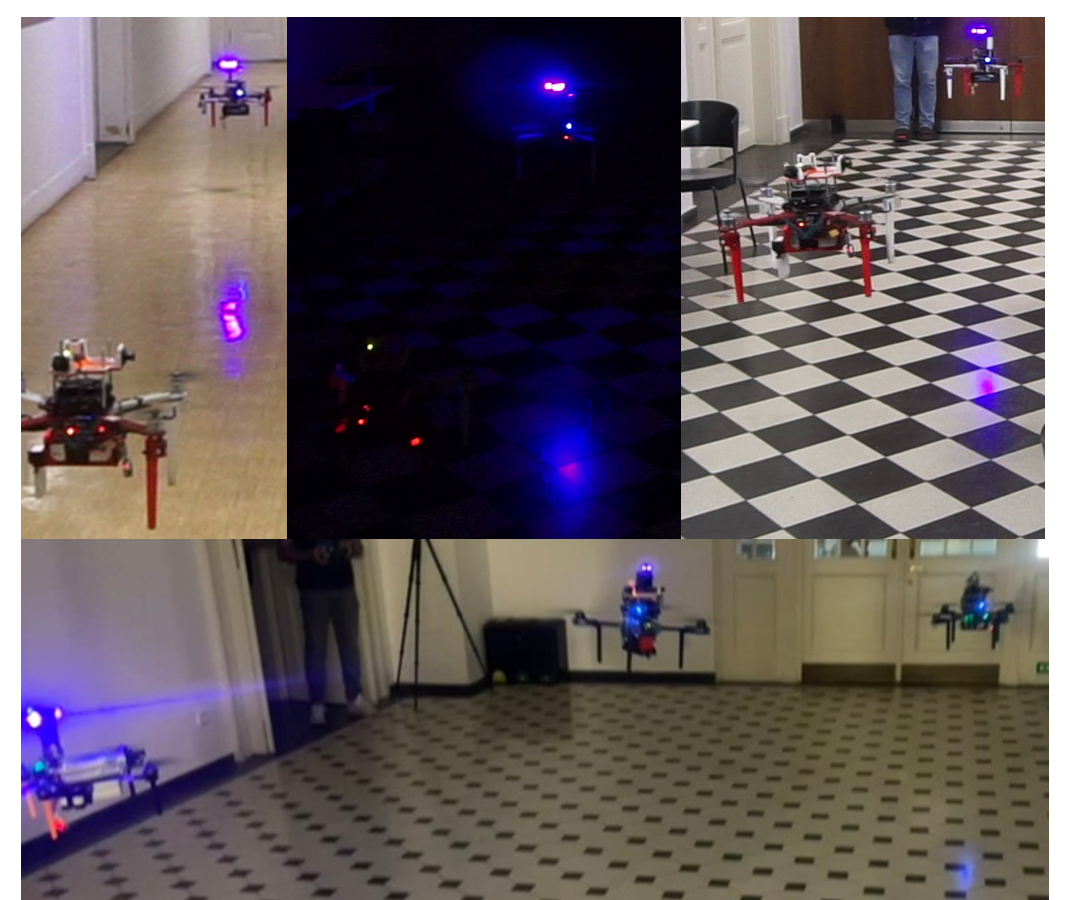}
   \caption{Indoor experiments on different floor types (\acs{PVC}, top left; tiles, top right) and under dark conditions (top center). Bottom: heterogeneous multi-\ac{UAV} experiment.}
   \label{fig:indoor_img_overview}
\end{figure}

The proposed system was conceived for \acp{UAV} operating over specular, near-planar surfaces such as bodies of water with wave heights on the order of centimeters, or flat indoor floors, where spurious lateral reflectors are rare.
When using \acp{LED} as active markers, reflection intensity follows the Fresnel equations~\cite{born1999principles} and varies with surface smoothness, ranging from specular (calm water, polished floors) to diffuse (rough surfaces), with most indoor environments falling between these extremes~\cite[p.~108]{hecht2017optics}.
In practice, outdoor surfaces are further affected by environmental disturbances such as wind and propeller downwash, making explicit surface modeling intractable.
While wall reflections are theoretically feasible, our experiments indicate that walls in urban indoor environments produce predominantly diffuse reflections, resulting in insufficient light intensity reaching the receiver (see Fig.~\ref{fig:intro_img} and top-left image in Fig.~\ref{fig:indoor_img_overview}).
Furthermore, reliable distance estimation from wall reflections would require knowledge of the wall orientation, necessitating a $2\mathrm{D}$ \ac{LIDAR} for vertically aligned walls or a $3\mathrm{D}$ \ac{LIDAR} for arbitrarily oriented surfaces.
Incorporating such sensors would add significant weight and (computational) cost, contradicting the lightweight sensor philosophy of our approach.
We therefore focus exclusively on ground and water reflections, for which a single dominant reflection is reliably obtained and the scalar distance to the surface is measured by a lightweight $1\mathrm{D}$ rangefinder, such as a single-beam \ac{LIDAR} or sonar, rather than the $2\mathrm{D}$ or $3\mathrm{D}$ \ac{LIDAR} required to recover full surface orientation.
The surface is consequently assumed level and near-planar. Local deviations from planarity are absorbed by the reflection cone model, and the position bias incurred by deviations from level is quantified in Sec.~\ref{subsec:tilt}.
For simplicity, the following derivation considers a single transmitting \ac{UAV}.

For a calibrated fisheye camera, the pixel coordinates of $p_d$ and $p_r$ can be converted into 3D unit bearing vectors using the function \emph{cam2world}, which is available in various computer vision libraries, such as \textit{OCamLib}~\cite{scaramuzzaFlexibleTechniqueAccurate2006a}.
In the following, the coordinate origin is defined at the body frame of the observer.
The $x$-axis aligns with the horizontal center of the camera and is parallel to the surface, the $y$-axis is also parallel to the surface, and the $z$-axis points upwards (Fig.~\ref{fig:method:tikz_v_view}).
To account for the roll ($\psi$) and pitch ($\phi$) angles of the observing aircraft, the 3D unit bearing vectors corresponding to $p_d$ and $p_r$ are transformed as follows:
\begin{align}
    \vec{\hat{b}}_{d} &= [\hat{x}_{d}, \hat{y}_{d}, \hat{z}_{d}]^{\top} = \mathbold{R}(-\psi)\mathbold{R}(-\phi)\textit{cam2world}(p_{d}), \label{eq:cam2world_pd}\\
    \vec{\hat{b}}_{r} &= [\hat{x}_{r}, \hat{y}_{r}, \hat{z}_{r}]^{\top} = \mathbold{R}(-\psi)\mathbold{R}(-\phi)\textit{cam2world}(p_{r}), \label{eq:cam2world_pr}
\end{align}
where $\mathbold{R}(-\psi)$ and $\mathbold{R}(-\phi)$ are the rotation matrices that compensate for the roll and pitch of the observing \acs{UAV}, respectively.
    While estimating the relative location from the intersection of the two bearing vectors is geometrically straightforward, surface irregularities and finite camera resolution introduce errors in both $\vec{\hat{b}}_d$ and $\vec{\hat{b}}_r$ that render this intersection ill-conditioned, producing large and unpredictable position errors.
As shown in Fig. \ref{fig:method:tikz_overview}, the center of the reflection, denoted by $\vec{r}$, is retrieved using the two angles $\beta_r$ and $\theta_r$, which are defined as follows: 
\begin{align}
        \beta_r&= \arctan{\frac{|\hat{z}_r|}{\hat{x}_r}},\label{eq:beta}\\
        \theta_r &= \arctan{\frac{\hat{y}_r}{\hat{x}_r}}. \label{eq:theta}
\end{align}
Thus, the center of the reflection, $\vec{r}$, is determined using equations \eqref{eq:cam2world_pr} to \eqref{eq:theta}, along with the height of the observer, $z_{o}$:
\begin{equation}
  \begin{aligned}
        \vec{r} &=[r_x \mkern9mu r_y \mkern9mu r_z]^{\top} \\&= \Big[
        |z_{o}|\frac{\hat{x}_r}{|\hat{z}_r|}\quad
        |z_{o}|\frac{\hat{y}_r}{|\hat{z}_r|}\quad
         - z_o\Big]^{\top}.\label{eq:r}
  \end{aligned}
\end{equation}

As illustrated in Fig. \ref{fig:intro_img} and \ref{fig:high_exposure}, the light emitted by the \ac{UAV} and its reflection on a surface appears elongated along the viewing axis.
Two factors cause this elongation: the geometric projection of the reflected light at shallow incidence angles and the scattering induced by surface roughness.
To represent the spread in the reflection, we construct an ellipse centered at $p_r$.
The minor and major axes are defined by the farthest pixels from $p_r$ in the vertical $(p_v)$ and horizontal $(p_h)$ directions, respectively.
These pixels are selected based on their intensity exceeding a predefined threshold $\sigma$ within a local bounding box around $p_r$.
To retrieve the vectors representing the minor and major axis of the ellipse, equations \eqref{eq:cam2world_pr} to \eqref{eq:theta} can be applied to $p_v$ and $p_h$ yielding:
\begin{align}
        \vec{r}_v &= \Big[|z_{o}| \frac{\hat{x}_v}{|\hat{z}_v|}\quad |z_o| \frac{\hat{y}_v}{|\hat{z}_v|}\quad-z_o\Big]^\top. \label{eq:r_v}\\
        \vec{r}_h &= \Big[|z_{o}| \frac{\hat{x}_h}{|\hat{z}_h|}\quad |z_o| \frac{\hat{y}_h}{|\hat{z}_h|}\quad-z_o\Big]^\top. \label{eq:r_h}
\end{align}

Using equations \eqref{eq:r} to \eqref{eq:r_h}, we construct an elliptical cone (red in Fig. \ref{fig:method:tikz_v_view}).
The apex of the cone is defined by the vector $\vec{a} = [0\,\,0\,\,-2z_0]^\top,$ and the half-apex angle of the cone along the vertical axis is $\alpha_r^{\text{max}} = \beta_v - \beta_r$, which corresponds to the major axis of the ellipse defined by the distance between $\vec{r}$ and $\vec{r}_v$ (Fig. \ref{fig:method:tikz_v_view}).
As shown in Fig. \ref{fig:method:tikz_h_view}, the horizontal spread of the cone along the horizontal axis (corresponding to the minor axis of the ellipse) defines the half-apex angle:
\begin{align}
        \upsilon_r^{\text{max}} = \arctan{\frac{||\vec{r}_h-\vec{r}||_{xy}}{||\vec{r}-\vec{a}||_{xy}}},
\end{align} 
in the $x$--$y$ plane.
Using a predefined maximum height $(z_{m})$, we define the end point of the elliptical cone by the vector:
\begin{align}
       \vec{p}_{r} = \begin{bmatrix}
        r_x + z_m\frac{\hat{x}_r}{|\hat{z}_r|}\\
        r_y + z_m\frac{\hat{y}_r}{|\hat{z}_r|} \\
        z_{m}
       \end{bmatrix}.\label{eq:P_r}
\end{align}
With equations \eqref{eq:beta} to \eqref{eq:P_r}, the following parametric equation defines a finite convex elliptical cone:
\begin{equation}
\begin{aligned}
    \vec{c}_r(\gamma_r, \alpha_r, \upsilon_r,l_r )= 
    \vec{a} &+ l_r\vec{\hat{d}}_r + l_r\vec{\hat{n}}_{1}\tan{\alpha_r}\cos{\gamma_r} \\
    &+ l_r\vec{\hat{n}}_{2}\tan{\upsilon_r}\sin{\gamma_r},\label{eq:parametric_cone_reflect}
\end{aligned}
\end{equation}
where:
\begin{align}
        &\gamma_r \in [0, 2\pi],
        &&\alpha_r \in [0,\alpha_r^{\text{max}}], 
        &\upsilon_r \in [0,\upsilon_r^{\text{max}}],\nonumber\\
        &l_r \in [0, ||\vec{p}_{r}-\vec{a}||],
        &&\vec{\hat{d}}_r = \frac{\vec{p}_{r}-\vec{a}}{||\vec{p}_{r}-\vec{a}||},\label{eq:reflection_cone_params}
\end{align}

and $\vec{\hat{n}}_{1}$ is a unit vector perpendicular to $\vec{\hat{d}}_r$, while $\vec{\hat{n}}_{2}$ is a unit vector perpendicular to both $\vec{\hat{d}}_r$ and $\vec{\hat{n}}_{1}$.

To address uncertainties arising from limited camera resolution and detection inaccuracies, we define two angles, $\alpha_d^{\text{max}}$ and $\upsilon_d^\text{max}$, which represent the vertical and horizontal uncertainties around $\vec{\hat{b}}_d$, respectively.
These uncertainties are used to create a finite convex elliptical cone around $\vec{\hat{b}}_d$, similar to the cone representation of the reflection vector and its associated uncertainty in \eqref{eq:parametric_cone_reflect}.
This cone is parameterized by:
\begin{equation}
\begin{aligned}
        \vec{c}_d(\gamma_d, \alpha_d, \upsilon_d,l_d) = l_{d} \vec{\hat{d}}_d&+ l_d\vec{\hat{n}}_{3} \tan{\alpha_d}\cos{\gamma_d}\\&+ l_d\vec{\hat{n}}_{4}\tan{\upsilon_d}\sin{\gamma_d},
         \label{eq:parametric_cone_direct}
\end{aligned}
\end{equation}
where:
\begin{align}
 &\gamma_d\in[0,2\pi], 
 &&\alpha_d \in [0,\alpha_d^{\text{max}}], 
 &\upsilon_d \in [0,\upsilon_d^{\text{max}}],\nonumber\\ 
 &l_d \in [0, ||\vec{p}_{d}||],
 && \vec{\hat{d}}_d = \frac{\vec{p_d}}{||\vec{p_d}||},
\end{align}
and $\vec{\hat{n}}_{3}$ is perpendicular to $\vec{\hat{b}}_{d}$, and $\vec{\hat{n}}_{4}$ is perpendicular to both $\vec{\hat{n}}_{3}$ and $\vec{\hat{b}}_d$.
The vector denoting the end point of the apex axis, depends on the predefined maximum extent in the $x$-direction $(x_{m})$ and is expressed as:
\begin{align}
        \vec{p}_{d} &=  \Big[x_{m} \quad x_{m} \frac{\hat{y}_d}{\hat{x}_d} \quad x_{m}\frac{\hat{z}_d}{\hat{x}_d}\Big]^{\top}.
\end{align}
The intersection area of the two cones, defined by equations \eqref{eq:parametric_cone_reflect} and \eqref{eq:parametric_cone_direct}, represents possible locations of the transmitting \ac{UAV}, resulting in a system of non-linear equations.
To estimate the location of the transmitting \ac{UAV} we employ a random sampling-based approach: we generate a set of vectors $\mathcal{X} = \{\vec{p}_i \sim U | i = 1,\hdots,n\}$ whose elements verify equation~\eqref{eq:parametric_cone_reflect}, and therefore ensuring they lie within the reflection cone.
The vectors in $\mathcal{X}$ are generated by sampling the parametric variables from uniform distributions:
\begin{align}
        \gamma'_r &\sim {U}(0, 2\pi),
        &\alpha'_r &\sim {U}(0,\alpha_r^{\text{max}}),\nonumber\\
        \upsilon'_r &\sim {U}(0,\upsilon_r^{\text{max}}), 
                    &l'_r &\sim {U}(l_{r}^{\text{min}}, ||\vec{p}_{r}-\vec{a}||),
\end{align}
where $l_{r}^{\text{min}} = ||\vec{a}-\vec{r}||$ ensures that only physically feasible points are sampled, i.e., points located above the surface and within the possible intersection volume.
To determine whether $\vec{p}_i \in \mathcal{X}$ lies within the cone defined by equation~\eqref{eq:parametric_cone_direct} and shown in blue in Fig. \ref{fig:method:tikz_overview}, we first compute its projection onto the cone axis:
\begin{align} \vec{{p}}_{i||} = \text{proj}_{\vec{\hat{d}}_d}\vec{p_i} = \big(\vec{p_i} \boldsymbol{\cdot} \vec{\hat{d}}_d\big) \vec{\hat{d}}_d, \end{align}
with $\boldsymbol{\cdot}$ denoting the scalar product, the two inequalities must hold:
\begin{align}
    \frac{(\vec{p}_i - \vec{{p}}_{i||})_y^2}{(||\vec{{p}}_{i||}||\tan{\upsilon_d^\text{max}})^2} + \frac{(\vec{p}_i - \vec{{p}}_{i||})_z^2}{(||\vec{{p}}_{i||}||\tan{\alpha_d^{\text{max}}})^2} &\leq 1 \label{eq:ellipse_constraint},\\
    0 \leq ||\vec{{p}}_{i||}|| &\leq l_d\label{eq:length_constrain}.
\end{align}
Inequality \eqref{eq:ellipse_constraint} ensures that the point lies within the elliptical cross-section of the cone, while inequality \eqref{eq:length_constrain} ensures that it lies within the extent of the cone along its axis.
If a point in $\mathcal{X}$ satisfies both conditions given by equations \eqref{eq:ellipse_constraint} and \eqref{eq:length_constrain}, it is included in the inlier set $\mathcal{X}_d$. 
The points in $\mathcal{X}_d$ are then used as observations for a particle filter with $n_p$ particles, which propagates according to a constant-velocity model and updates particle weights using a robust Cauchy kernel.
    The particles are initialized using a uniform distribution over the $3\mathrm{D}$ space observable by the camera.
Resampling is triggered when the \ac{ESS} falls below a predefined threshold, set to half the total number of particles.
The resulting weighted particle cloud is then used to compute the final position estimate of the transmitting \ac{UAV}.
\section{Experiments}\label{sec:eval}
\begin{figure}[ht!]
    \centering
    \captionsetup[subfigure]{labelformat=empty}
    \begin{subfigure}[b]{0.45\textwidth}
       \def\svgwidth{1\textwidth}
       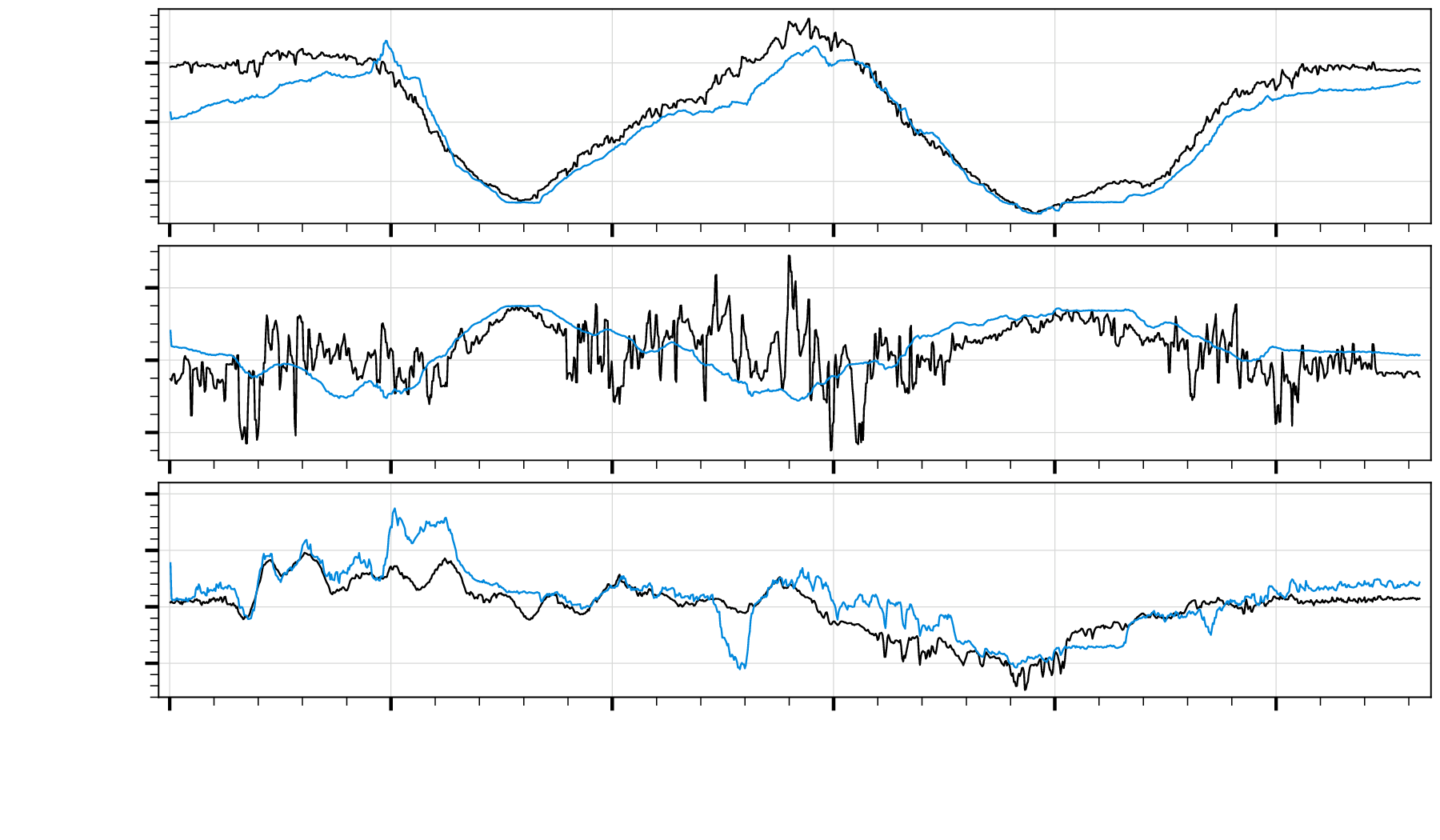
      \caption{}
      \label{fig:indoor_exp_1}
      \vspace{-10pt}
    \end{subfigure}
    \begin{subfigure}[b]{0.45\textwidth}
       \def\svgwidth{1\textwidth}
       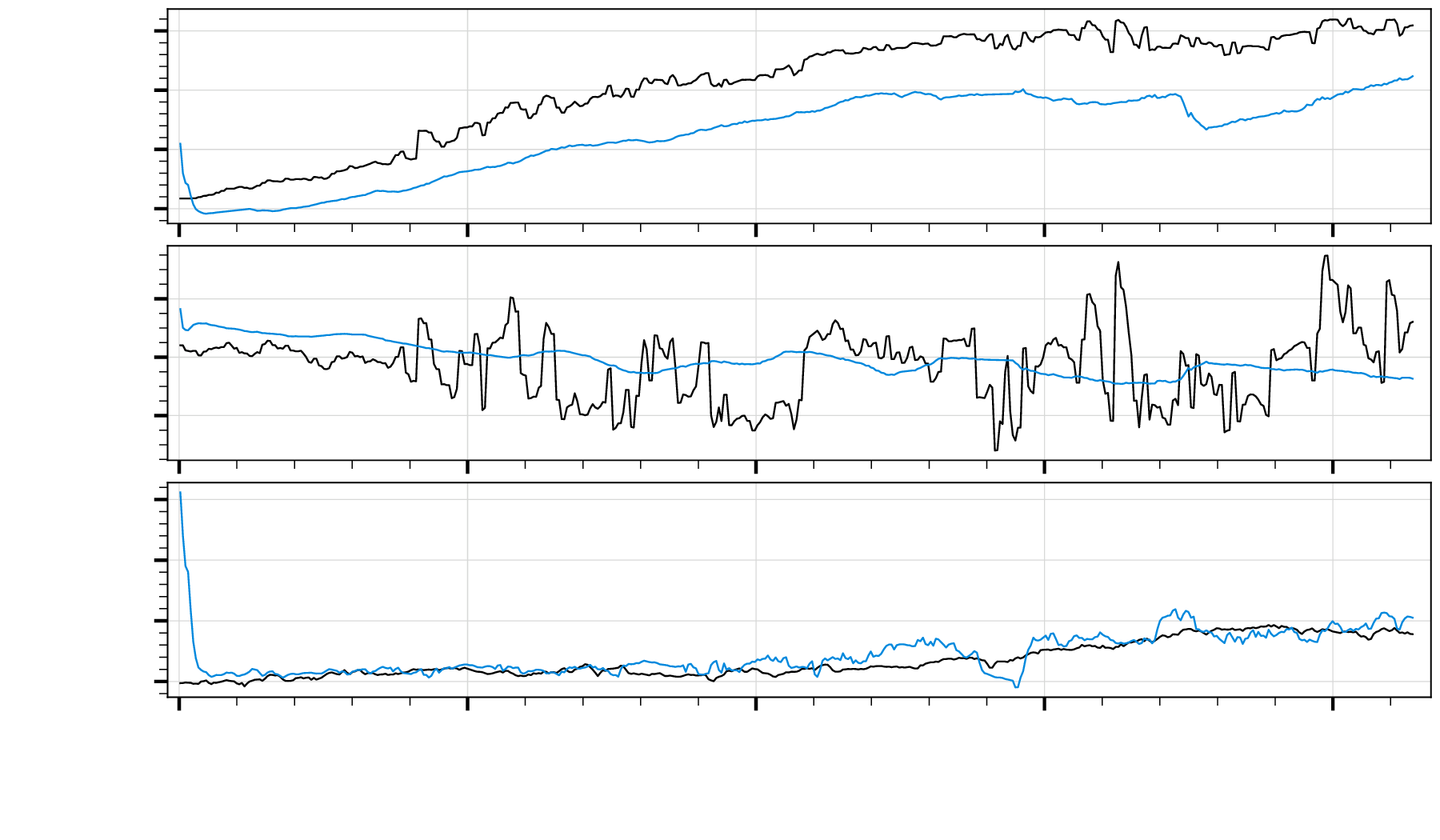
      \caption{}
      \label{fig:indoor_exp_2}
      \vspace{-10pt}
    \end{subfigure}
    \begin{subfigure}[b]{0.45\textwidth}
       \def\svgwidth{1\textwidth}
       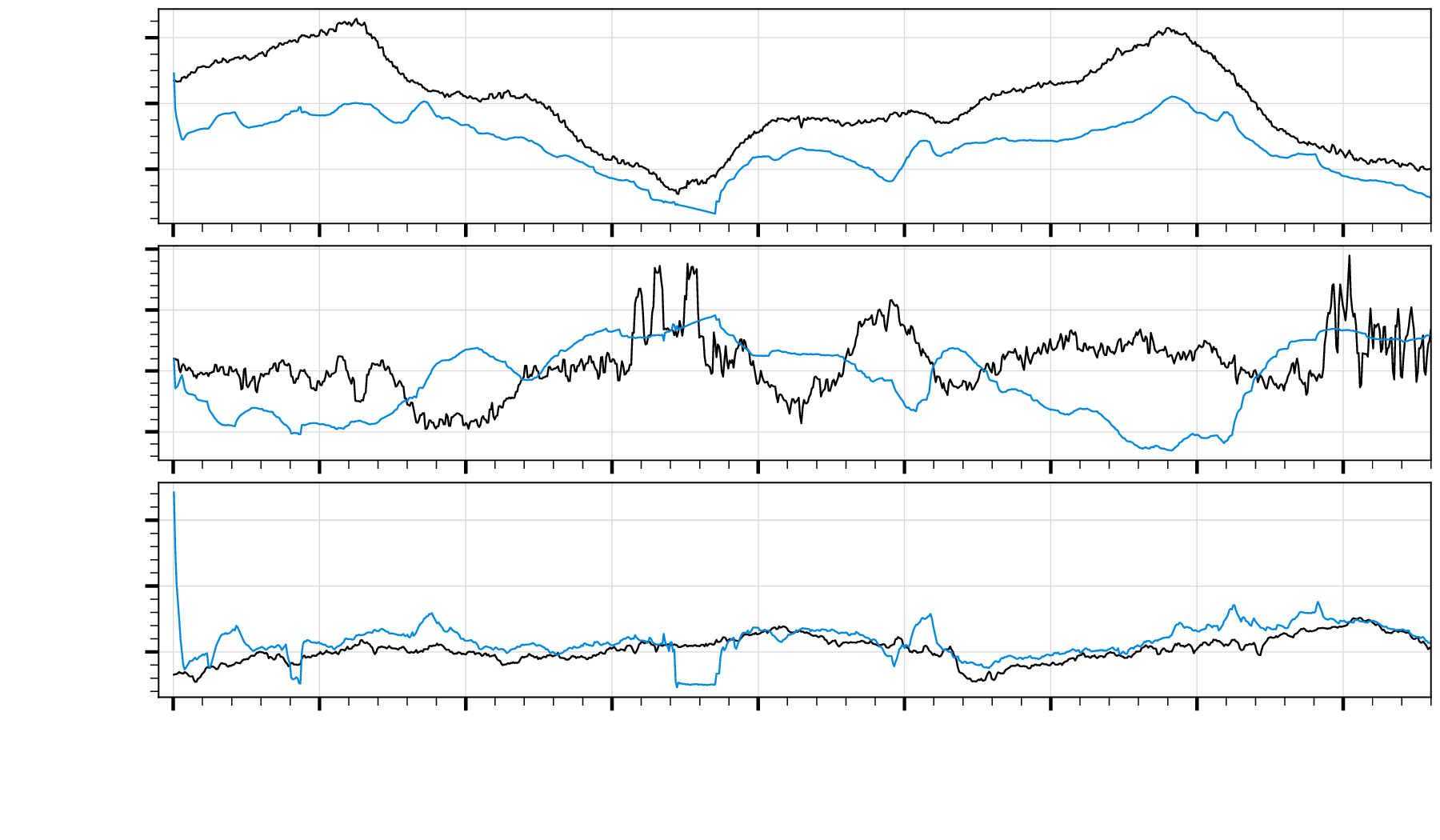
      \caption{}
      \label{fig:indoor_exp_3}
      \vspace{-10pt}
    \end{subfigure}
    \begin{subfigure}[b]{0.45\textwidth}
       \def\svgwidth{1\textwidth}
       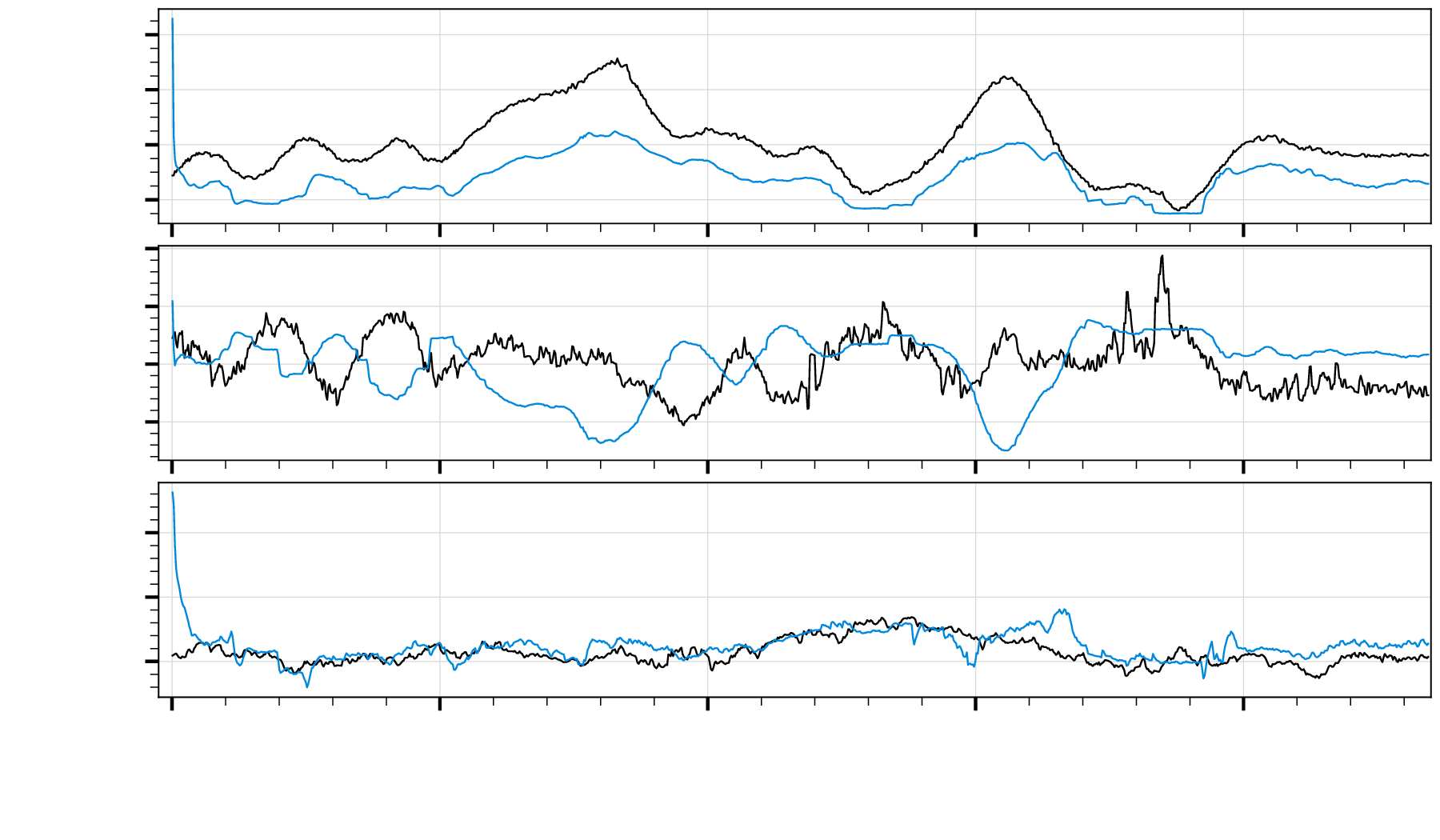
        \caption{}
      \label{fig:indoor_exp_4}
    \end{subfigure}
      \vspace{-10pt}
    \caption{Relative position estimates obtained using our approach (blue) compared to the relative position ground truth from \ac{UWB} (black) during indoor experiments 1--4 (from top to bottom).}
    \label{fig:indoor_exp}
    \vspace{-10pt}
\end{figure}
\begingroup
\setlength{\tabcolsep}{3pt}
\renewcommand{\arraystretch}{1.1}
\begin{table}[b!]
  \centering
  \footnotesize
  \setlength{\tabcolsep}{3pt}
  \begin{tabular}{l | c c c c c c c c c}
    \hline
    \textbf{Experiment} & $n$ & $n_p$ & $\alpha_{r}^{\max}$ & $\alpha_{d}^{\max}$ & $\upsilon_{d}^{\max}$ & $x_m$ & $z_m$ & $\sigma$ & $t_{\text{exp}}$ \\
     & & & [$^\circ$] & [$^\circ$] & [$^\circ$] & [m] & [m] & & [ms] \\
    \hline
    Indoor 1--4  & 500  & 1500 & 3 & 3 & 3 & 15      & 2.5 & 30 & 20 \\
    Multi--UAV   & 500  & 1500 & 3 & 3 & 3 & 10      & 2.5 & 30 & 10 \\
    Outdoor 1--3 & 1000 & 1000 & 4 & 3 & 3 & 30--40  & 10  & 40 & 2--3 \\
    \hline
  \end{tabular}
  \caption{Parameters for indoor and outdoor experiments.}
  \label{tab:param_overview}
\end{table}
\endgroup
\begin{figure}[ht]
    \captionsetup[subfigure]{labelformat=empty}
    \centering
    \begin{subfigure}[b]{0.45\textwidth}
       \def\svgwidth{1\textwidth}
       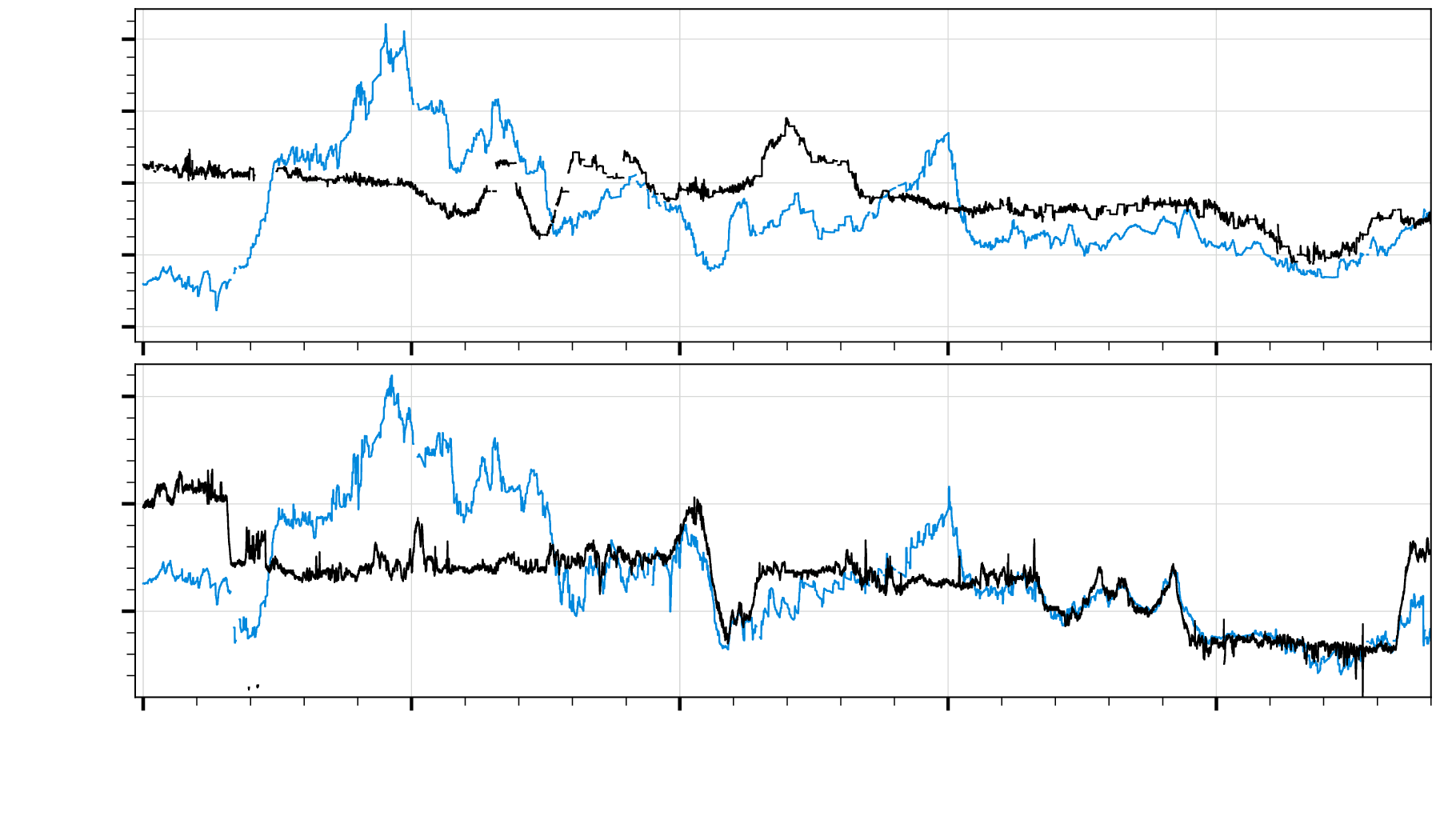
      \caption{}
    \label{fig:eval:online_id1}
    \end{subfigure}
    \begin{subfigure}[b]{0.45\textwidth}
      \vspace{-10pt}
       \def\svgwidth{1\textwidth}
       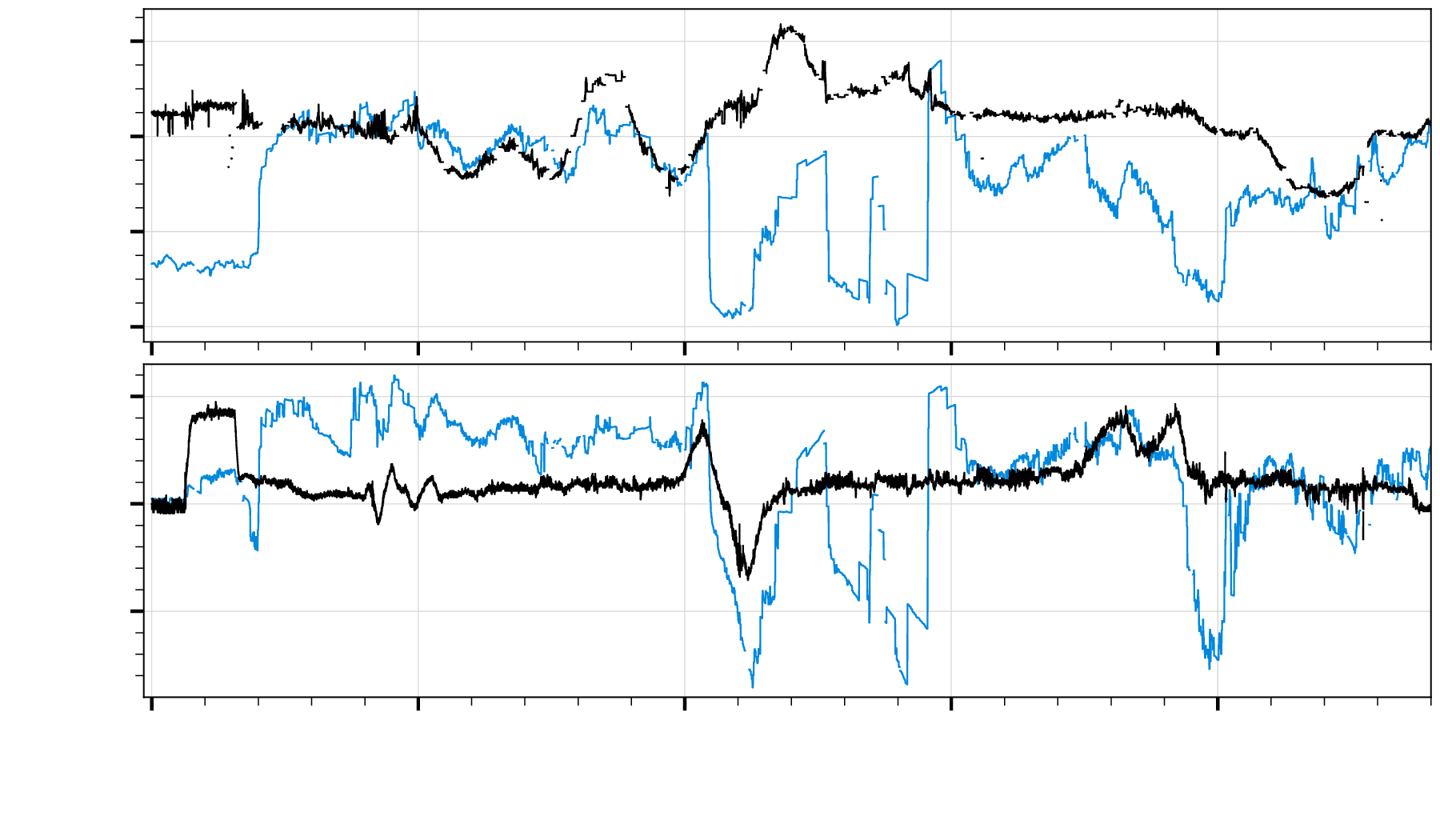
       \caption{\label{fig:eval:online_id4}}
    \end{subfigure}
      \vspace{-10pt}
      \caption{Relative position estimates obtained using our approach (blue) and the ground truth from \ac{UWB} (black) for the \ac{UAV} (top) and \ac{MAV} (bottom) during the multi-\ac{UAV} experiment.}
    \label{fig:eval:online}
\end{figure}
We evaluated the approach in both indoor and outdoor experiments, comparing results against the state-of-the-art \green{vision-based active-marker} method.
Unless stated otherwise, processing was performed post-flight on a local machine (\emph{Intel i7-8550U}, \SI{1.8}{\giga\hertz}).
\green{Onboard execution} is demonstrated in the multi-\ac{UAV} experiment (Sec.~\ref{sec:multi-uav}).
Ground truth was provided by \ac{UWB} indoors and \ac{RTK} outdoors.
In the following, we distinguish between two platform types. 
The \ac{UAV} refers to platforms based on the \emph{F450} or \emph{Holybro X500} frame~\cite{hertMRSDroneModular2023}, equipped with an \emph{Intel NUC 10 i7FNK} (6 cores, up to \SI{4.7}{\giga\hertz}), while the \ac{MAV} refers to a custom \SI{27}{\centi\meter} platform equipped with a \emph{Raspberry Pi}~5 (4 cores, up to \SI{2.4}{\giga\hertz}, \SI{8}{\giga\byte} RAM).
Both platforms carried a camera with a fisheye lens and \ac{UV} bandpass filter. 
The \acp{UAV} used a $752\times480$ BlueFOX camera operating at \qtyrange{10}{30}{\hertz}, while the \ac{MAV} carried an ArduCam operating at $640\times400$ and \SI{10}{\hertz}.
While a downward-facing \ac{LIDAR}-based rangefinder was used for height estimation indoors, \ac{RTK} was employed outdoors.
Both sensors relayed measurements to the onboard PC via MAVLink, with MAVROS providing clock synchronization and interpolating height measurements to camera timestamps.
For outdoor experiments, we mounted \ac{UV}-\acp{LED} on the arms of the transmitter \ac{UAV}, with all \acp{LED} emitting an identical signal, consistent with the \ac{UVDAR} setup in~\cite{walterUVDARSystemVisual2019}.
\green{During the indoor and outdoor experiments the observing \ac{UAV} was hovering with roll and pitch angles below \SI{0.06}{\radian}, so the camera frame was used as the gravity-aligned frame.}
\begingroup
\setlength{\tabcolsep}{4pt}
\renewcommand{\arraystretch}{1.05} 
\begin{table}[b]
  \centering
  \footnotesize
  \begin{tabular}{ l    c   c    c   c }
      \hline
       & \textbf{1. \acs{PVC} (d)} & \textbf{2. \acs{PVC} (n)} & \textbf{3. Tiles (n)} & \textbf{4. Tiles (n)}  \\
    \hline    
      \acs{MAE}$(x)$ [m]& $1.29 \pm 0.67$ & $2.12 \pm 0.70$ & $1.30 \pm 0.61$ & $1.22 \pm 0.60$\\
      \acs{MAE}$(y)$ [m]& $0.65 \pm 0.52$ & $1.09 \pm 0.74 $& $0.68 \pm 0.42$ & $0.61 \pm 0.41$\\
      \acs{MAE}$(z)$ [m]& $0.12 \pm 0.09$ & $0.08 \pm 0.12 $& $0.10 \pm 0.09$ & $0.08 \pm 0.09$\\ 
      \hline
  \end{tabular}
  \caption{\acf{MAE} in the $x$-, $y$-, and $z$-axes for indoor experiments conducted under daylight (d) and nighttime (n) conditions on different floor types. Values are reported as \ac{MAE} $\pm$ one standard deviation.}
  \label{tab:eval:indoor}
\end{table}
\endgroup
Tab.~\ref{tab:param_overview} summarizes the parameters used in the experiments. 
Exposure time settings were adapted depending on the environment (indoor vs. outdoor), with the local binary threshold ($\sigma$) adjusted accordingly.
Additionally, the number of samples in the reflection cone $n$ was reduced in the multi-\ac{UAV} experiment, to ensure \green{onboard deployability} on the \ac{MAV}.
\subsection{Indoor}
Indoor experiments were conducted in two environments: a standard office hallway with a \ac{PVC} floor and a corridor with a tiled surface (Fig.~\ref{fig:indoor_img_overview}).
To highlight platform independency, a single circular \ac{LED} array was mounted on top of each \ac{UAV} (Fig.~\ref{fig:indoor_img_overview}).
The \ac{UVDAR} system could not be evaluated in this setup, as the \ac{LED} array on the transmitting \ac{UAV} had minimal spacing between \acp{LED}, while the \ac{UVDAR} system requires at least two spatially separated markers.
Height estimation on the observer was performed using a downward-facing \ac{LIDAR}-based rangefinder.
The experiments are divided into two-\ac{UAV} and multi-\ac{UAV} configurations.
\subsubsection{Two-\ac{UAV} Experiment}
Fig.~\ref{fig:indoor_exp} shows the resulting relative position estimates compared to \ac{UWB} ground truth.
Tab.~\ref{tab:eval:indoor} summarizes the \acfp{MAE} and standard deviations per axis across all experiments.
Consistently, the largest \ac{MAE} was observed along the $x$-axis (optical axis), which is expected as depth estimation from a monocular camera is inherently less constrained than lateral estimation.
The method performed reliably across both floor types and under both daylight and nighttime conditions, demonstrating robustness to varying surface properties and lighting.
\subsubsection{Multi-\ac{UAV} Experiment}\label{sec:multi-uav}
The system ran onboard all three \ac{UAV} platforms, including the \ac{MAV} equipped with a \emph{Raspberry Pi}~5, demonstrating deployability on low-computational-power hardware.
The clustering radius was set to $r_c = 10$ pixels to account for the higher exposure settings required for reflection detection.
Fig.~\ref{fig:eval:online} shows the estimated relative position, decomposed into the planar Euclidean distance ($d_{xy}$) and vertical ($z$) components, from one UAV to another \ac{UAV} and to one \ac{MAV}.
The method achieves an \ac{MAE} of $1.50\pm\SI{1.44}{\meter}$ in $d_{xy}$ and $0.49 \pm \SI{0.36}{\meter}$ in $z$ for the \ac{UAV}, and $1.25 \pm \SI{0.93}{\meter}$ in $d_{xy}$ and $0.36 \pm \SI{0.37}{\meter}$ in $z$ for the \ac{MAV}. 
\green{The system reliably estimated the relative position of the \ac{UAV} and \ac{MAV} outside the interference interval from \SI{105}{\second} to \SI{155}{\second}, during which the \ac{MAV} was deliberately positioned in front of the observer to induce signal interference.}
Full position recovery requires both a reliable blinking-sequence extraction and the direct bearing $\vec{p}_d$ under \ac{LoS}, both of which degrade in this configuration.
Relative bearing alone remains obtainable but does not determine range.
\begin{figure}[t]
      \centering
        {\scalebox{.67}{\input{outdoor_exp_pond_2}}}
        \caption{Outdoor experiment: (Left) Observing \acs{UAV} (white), transmitting \acs{UAV} with active markers (blue), and surface reflections (red). (Right) Onboard \ac{UV}-filtered image showing the detected marker centroid (blue) and reflection ellipse (red).
}
        \label{fig:outdoor_experiment:uv_img}
\end{figure}
The per-timestamp computational complexity is dominated by the cone sampling and particle filter update, yielding $\mathcal{O}(n \cdot n_p)$ in the worst case, where $n$ is the number of samples drawn in the reflection cone and $n_p$ the number of particles.
In practice, the cone sampling accounts for approximately \SI{80}{\percent} of the per-frame latency, while the particle filter update remains below \SI{3}{\milli\second}.
The median estimator latency was \SI{78}{\milli\second} on the \emph{Intel NUC} and \SI{97}{\milli\second} on the \emph{Raspberry Pi 5}, exceeding the \SI{100}{\milli\second} camera period for \SI{19.6}{\percent} and \SI{45.4}{\percent} of frames, respectively.
As the estimator always processes the newest frame, an overrun lowers the update rate rather than accumulating latency, yielding mean rates of \SI{8.4}{\hertz} and \SI{6.8}{\hertz} at the accuracy reported above.
\begingroup
\setlength{\tabcolsep}{4.pt}
\renewcommand{\arraystretch}{1.01} 
\begin{table*}[ht!]
  \footnotesize
\vspace{5pt}
  \center
  \begin{tabular}{ l l | c c c | c  c c | c  c  c  }
      \hline    
      & & \multicolumn{3}{ c|}{\textbf{X}} & \multicolumn{3}{ c| }{\textbf{Y}}& \multicolumn{3}{ c  }{\textbf{Z}} \\ 
      & & \acs{MAE}[m] & \emph{p}-value & \SI{95}{\percent} \acs{CI} &  \acs{MAE}[m] & \emph{p}-value & \SI{95}{\percent} \acs{CI} & \acs{MAE}[m] & \emph{p}-value & \SI{95}{\percent} \acs{CI} \\
      \hline
      \multirow{2}{*}{\textbf{1. Outdoor}}  & Ours & $3.50 \pm 1.86$  & \multirow{2}{*}{\num{1.9e-8}}& \multirow{2}{*}{$[-2.62, -1.61]$}& $1.18\pm 0.80$ &\multirow{2}{*}{\num{0.015}}& \multirow{2}{*}{$[-0.26, -0.07]$}& $1.04\pm 1.57$& \multirow{2}{*}{\num{0.917}}& \multirow{2}{*}{$[0.15, 0.42]$}\\

                           & \acs{UVDAR} & $5.58\pm 5.77$ &  & & $1.37\pm 1.09$ & & &$0.78\pm 0.83$&&\\
      \hline
  \multirow{2}{*}{\textbf{2. Outdoor}} & Ours & $2.42\pm 1.50$ &\multirow{2}{*}{\num{2.0e-5}} &  \multirow{2}{*}{$[-0.75, -0.29]$}& $1.73\pm 0.95$ &\multirow{2}{*}{\num{0.071}}& \multirow{2}{*}{$[-0.16, 0.02]$} & $0.70\pm 0.47$&\multirow{2}{*}{\num{0.009}} &  \multirow{2}{*}{$[-0.01, 0.07]$}\\
                           & {\acs{UVDAR}} & $2.97\pm 2.14$ & &&$1.76\pm 0.97$ &&& $0.67\pm 0.44$&&\\
      \hline
  \multirow{2}{*}{\textbf{3. Outdoor}} & Ours & $5.01 \pm 4.15$ &\multirow{2}{*}{\num{2.2e-11}}& \multirow{2}{*}{$[-1.95, -1.18]$}& $3.29 \pm 1.58$ & \multirow{2}{*}{\num{0.000}}&\multirow{2}{*}{$[-4.89, -4.19]$}&$0.98 \pm 0.85$&\multirow{2}{*}{\num{0.793}}&\multirow{2}{*}{$[-0.04, 0.11]$}\\
                                       & \acs{UVDAR} & $7.00 \pm 6.03$ &&& $7.96 \pm 5.60$ &&& $0.97\pm 0.67$&&\\
      \hline
  \end{tabular}
    \caption{\acf{MAE} and one standard deviation along the $x$-, $y$-, and $z$-axes, together with the \emph{Wilcoxon p-values} and \SI{95}{\percent} \acf{CI} of the paired differences, comparing our method with \ac{UVDAR} in outdoor experiments.}
  \label{tab:eval:outdoor}
\end{table*}
\endgroup
\begingroup
\setlength{\tabcolsep}{3.5pt}
\renewcommand{\arraystretch}{1.05} 
\begin{table}[b]
    \centering
    \footnotesize
    \begin{tabular}{ l | c c | c c | c c }
       \hline
        & \multicolumn{2}{c|}{$x$} & \multicolumn{2}{c|}{$y$} & \multicolumn{2}{c}{$z$} \\
        & \SI{-15}{\percent} & $+$\SI{15}{\percent} & \SI{-15}{\percent} & $+$\SI{15}{\percent} & \SI{-15}{\percent} & $+$\SI{15}{\percent} \\
       \hline
        \textbf{1. Outdoor} & $+0.33$ & $-0.23$ & $-0.03$ & $-0.01$ & $+0.02$ & $-0.04$ \\
        \textbf{2. Outdoor} & $+0.07$ & $+0.14$ & $+0.39$ & $-0.16$ & $-0.05$ & $+0.09$ \\
        \textbf{3. Outdoor} & $+0.92$ & $+1.22$ & $+0.10$ & $-0.55$ & $+0.30$ & $+0.15$ \\
       \hline
    \end{tabular}
    \caption{Sensitivity analysis: change in \ac{MAE} [m] under systematic bias in the height measurement of the observer \ac{UAV}, relative to the unbiased results in Tab.~\ref{tab:eval:outdoor}.}
    \label{tab:height}
\end{table}
\endgroup
\subsection{Outdoor}
\begingroup
\setlength{\tabcolsep}{3.5pt}
\renewcommand{\arraystretch}{1.0}
\begin{table*}[t]
\centering\footnotesize
\begin{tabular}{l c c c c c}
\hline
 & \textbf{Ours} & UVDAR~\cite{walterUVDARSystemVisual2019} & CREPES~\cite{xunCREPESCooperativeRElative2023a} & {Swarm-LIO2}~\cite{SwarmLIO2} & CNN~\cite{schillingVisionBasedDroneFlocking2021} \\
\hline
Hardware                          & Cam, LED, Rng               & Cam, LED              & Cam, LED, \acs{UWB}, IMU  & \acs{LIDAR}, IMU        & Cam, GPU \\
Mass / power [g / W]$^{1}$ & $62$ / $6.5$             & $40$ / $5.9$        & $130^{\star}$/ $\leq\!4^{\star}$ & $265$ / $6.5$        & $200^{\star}$ / $17^{\star}$ \\
Output rate [Hz]           & $10$                     & $60$                & $100$                     & $30$                 & $5$ \\
  Error @ range [m]$^{2}$ & \green{MAE$_x$ $2.4$--$5.0$ @ $\leq\!40$} & \green{MAE$_x$ $3.0$--$7.0$ @ $\leq\!40$}
  & P25--P75 $0.10$--$0.16$ @ $\leq\!27.5$ & std.~$0.05$
  & P25--P75 $0.3$--$0.7$ @ $6$ \\

Approx.\ cost [\$]         & $570$                    & $450$               & $500^{\star}$             & $700^{\star}$        & $500^{\star}$ \\
Direct ID                  & \yes                     & \yes                & \yes                      & \no                  & \no \\
\acs{RF}-free              & \yes                     & \yes                & \no                       & \yes                 & \yes \\
Lighting-robust            & \yes                     & \yes                & \partmark                 & \yes                 & \no \\
\hline
\end{tabular}
\caption{Comparison of onboard relative localization approaches.
Symbols: \yes~yes, \partmark~partial, \no~no.
$^{1}$Sensing hardware only, onboard compute excluded.
$^{2}$Position error as reported in the original publications, following the convention of each work. \green{MAE$_x$: mean absolute error along the optical axis}, P25--P75: interquartile range, std.: swarm-state consistency standard deviation reported by the authors, as mutual-distance ground truth was unavailable. \green{The metrics are not directly comparable.}
$^{\star}$Estimated from datasheets or manufacturer specifications.}
\label{tab:eval:comp}
\end{table*}
\endgroup
\begin{figure}[t]
    \captionsetup[subfigure]{labelformat=empty}
    \centering
    \begin{subfigure}[b]{0.45\textwidth}
       \def\svgwidth{1\textwidth}
       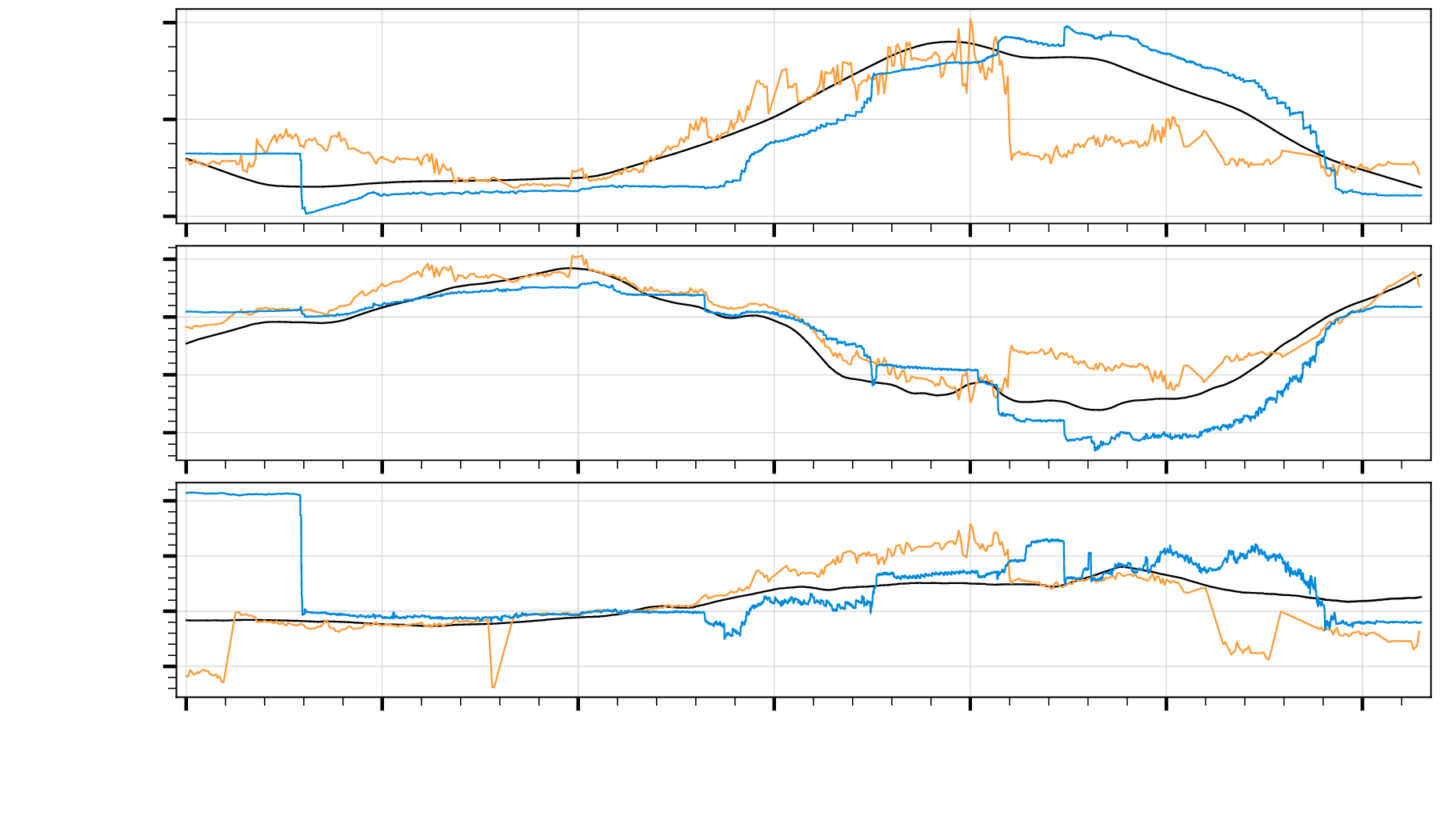
      \caption{}
    \label{fig:exp_outdoor_1}
    \end{subfigure}
    \begin{subfigure}[b]{0.45\textwidth}
      \vspace{-12pt}
       \def\svgwidth{1\textwidth}
       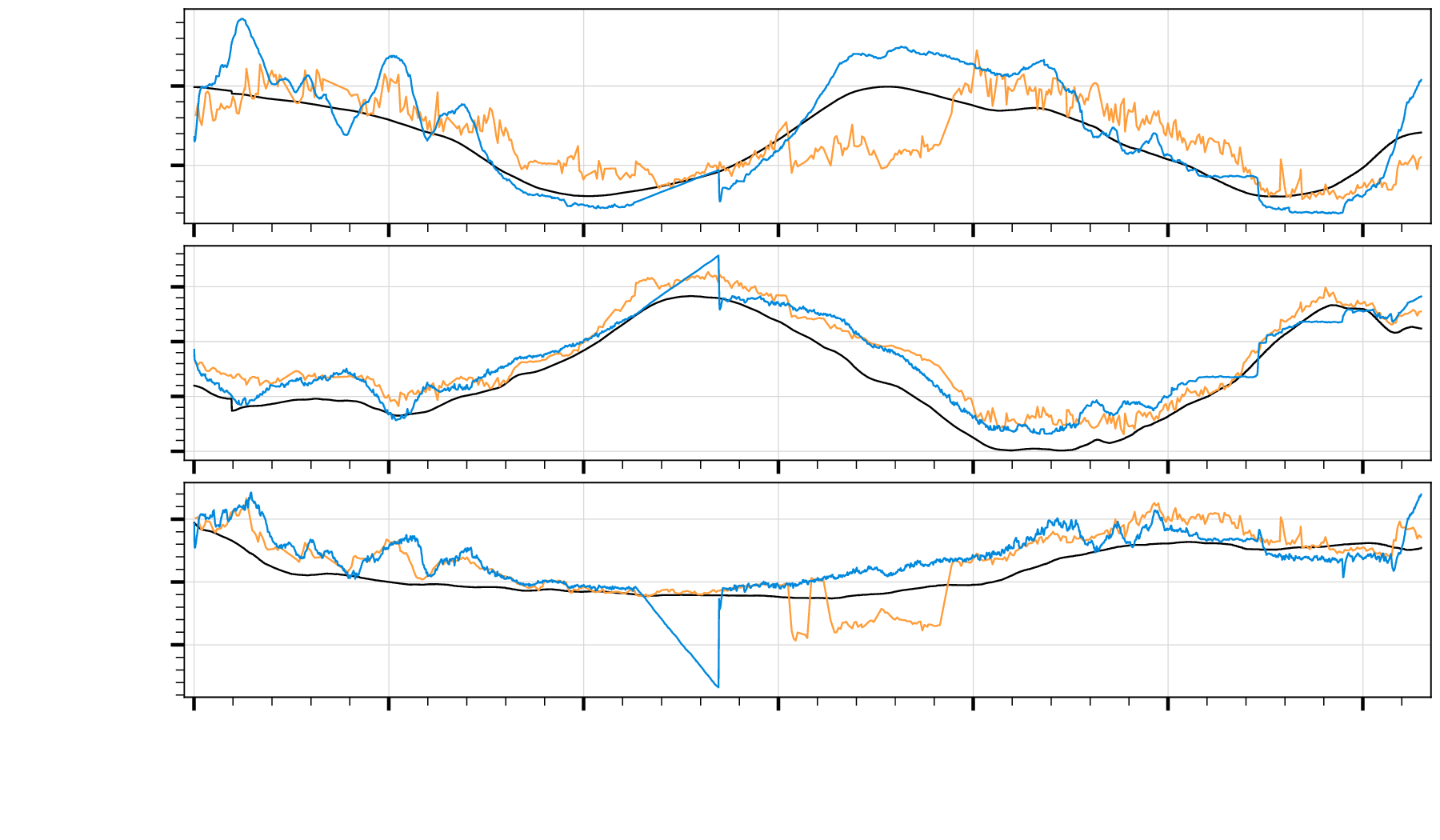
       \caption{\label{fig:exp_outdoor_2}}
    \end{subfigure}
    \begin{subfigure}[b]{0.45\textwidth}       
      \vspace{-12pt}
       \def\svgwidth{1\textwidth}
       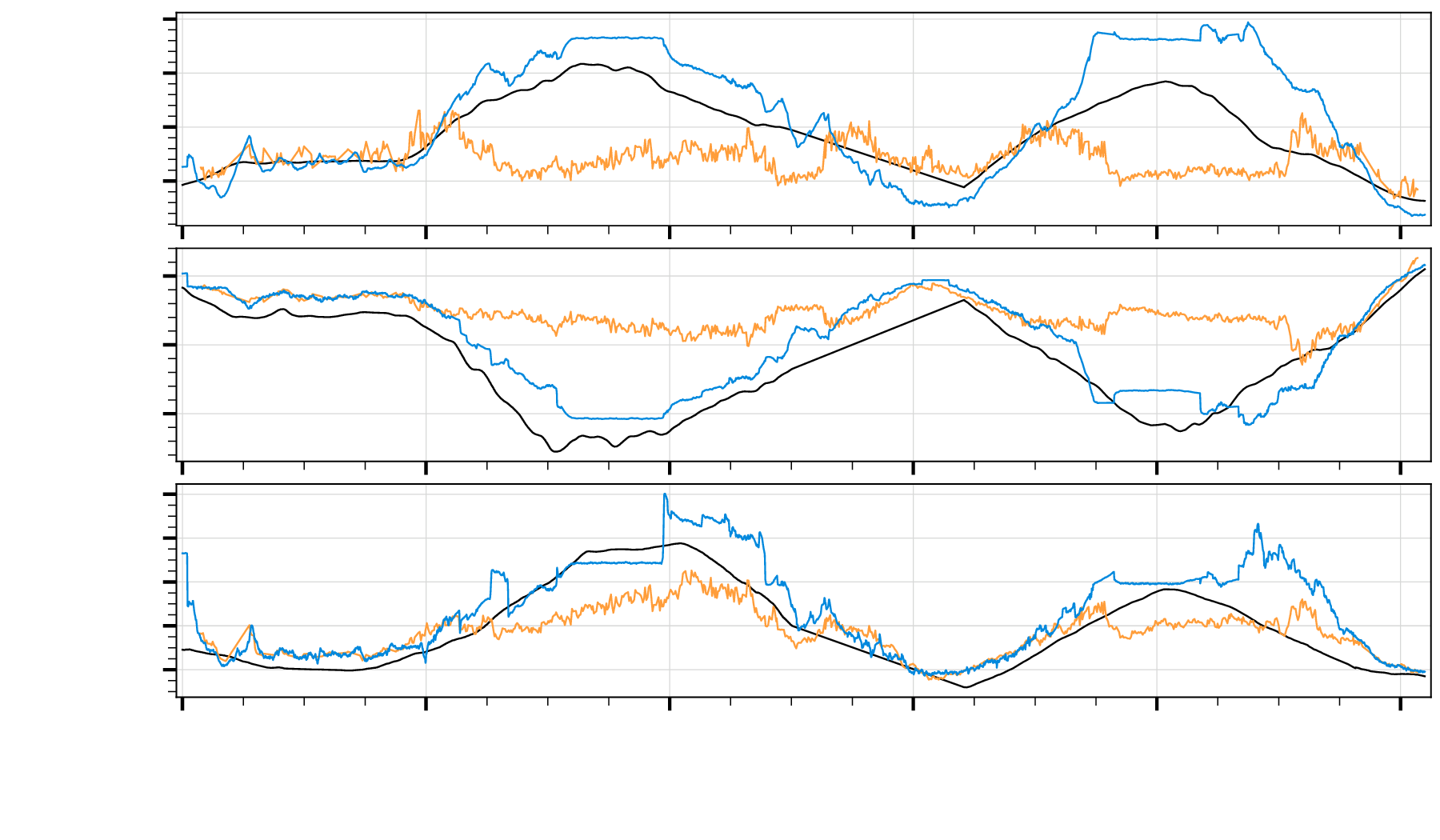
       \caption{\label{fig:exp_outdoor_3}}
    \end{subfigure}
      \vspace{-15pt}
    \caption{Relative position estimates obtained using our approach (blue), \ac{UVDAR} (orange), and the ground truth from \ac{RTK} (black) for outdoor experiments~1--3 (from top to bottom).}
    \label{fig:exp_outdoor}
\end{figure}
Fig.~\ref{fig:outdoor_experiment:uv_img} provides an overview of the outdoor experimental setup.
Multiple outdoor experiments with varying exposure time settings were conducted (Tab.~\ref{tab:param_overview}).
For example, in the experiment shown in Fig.~\ref{fig:high_exposure}, automatic exposure and gain control were used.
We observed that with a maximum exposure time of \SI{3}{\milli\second}, the camera maintained a stable frame rate (essential for reliable signal extraction) while reflections remained clearly visible at distances exceeding \SI{20}{\meter} on the water surface.

Fig.~\ref{fig:exp_outdoor} shows the three most relevant outdoor flights, comparing the relative positions obtained from our method, \ac{UVDAR}, and the ground truth.
Our reflection-based approach maintained high accuracy at distances beyond \SI{30}{\meter}, whereas \ac{UVDAR} struggled to provide reliable estimates past \SI{20}{\meter}.
This highlights the advantage of our approach, which operates without prior knowledge of the \ac{UAV} size or the configuration of its light sources.

Tab.~\ref{tab:eval:outdoor} summarizes the \acp{MAE} and standard deviations for each axis across all experiments, along with the \emph{Wilcoxon p-values} and \SI{95}{\percent} \ac{CI} of the paired differences.
Along the $z$-axis, \ac{UVDAR} achieved slightly lower \ac{MAE} values (\qtyrange{1}{25}{\percent}).
However, the differences were not statistically significant.
Along the $y$-axis, our method reduced the \ac{MAE} by \SI{13.9}{\percent} and \SI{58.7}{\percent} in experiments~1 and~3 (\emph{p-value} $< 0.05$, \SI{95}{\percent} \ac{CI} fully below zero), whereas no statistically significant difference was observed in experiment~2.
Along the $x$-axis (optical axis), our method reduced the \ac{MAE} by $18.5$--\SI{37.3}{\percent} compared to \ac{UVDAR}, with the difference being statistically significant (\emph{p-value} $< 0.05$, \SI{95}{\percent} \ac{CI} fully below zero).
\begin{figure}[b]
\centering
\def\svgwidth{0.48\textwidth}
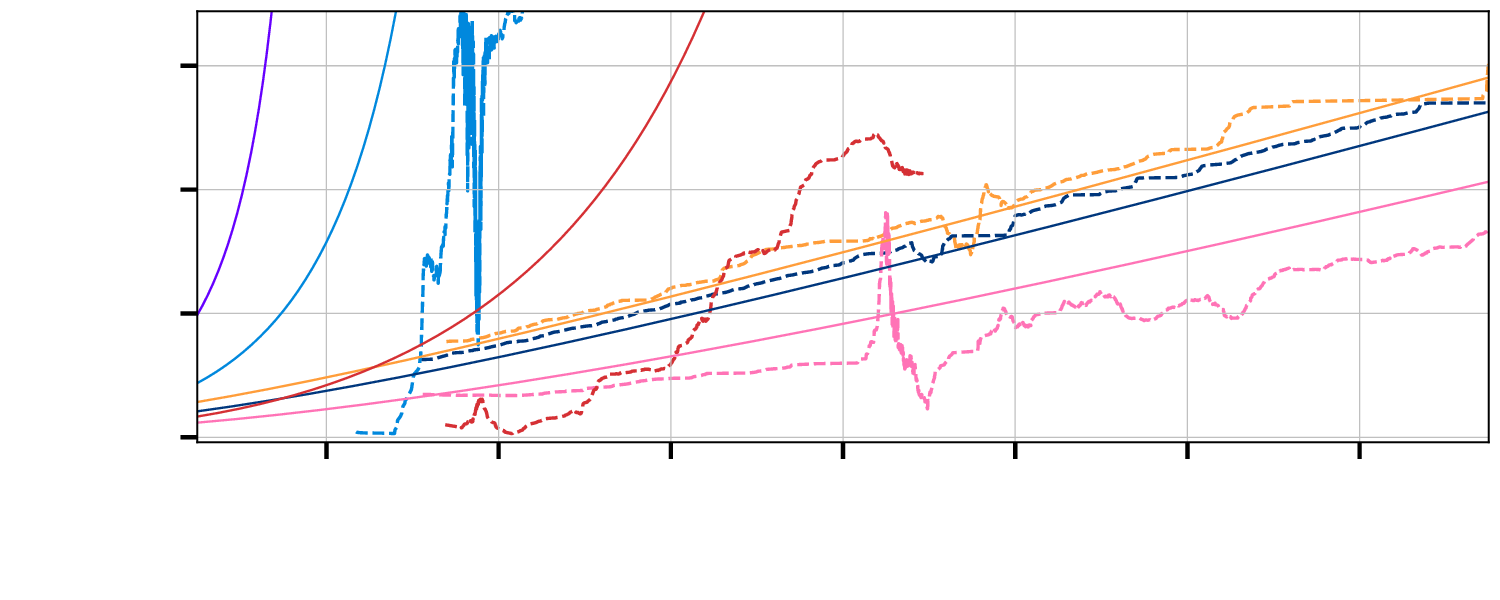

\caption{
  Euclidean localization error vs.\ range for surface tilt angles
\textcolor{ptmorange}{$15^\circ$},
\textcolor{ptmblue}{$10^\circ$},
\textcolor{ptmpink}{$5^\circ$},
\textcolor{ptmred}{$-5^\circ$},
\textcolor{ptmlightblue}{$-10^\circ$},
\textcolor{ptmpurple}{$-15^\circ$}.
Solid lines show the theoretical prediction, dashed lines the measured moving average.
No measurement is shown for $\tau = -15^\circ$, where the geometry is singular within the evaluated range and the method produced no output.
}
\label{fig:exp:tilt_surface}
\end{figure}
\subsection{Robustness to Model Assumptions}\label{subsec:tilt}
We evaluate robustness to two violations of the idealized model.
The first is error in the height estimation by the observer.
The second is tilt of the reflective surface.
Height error is characterized on the outdoor water dataset, where barometric drift is a realistic failure mode.
Surface tilt is characterized indoors, the setting where floor slope is the relevant deviation from the level-surface assumption.

To assess sensitivity to height error (e.g., from barometric sensors), we introduced systematic biases of up to $\pm\SI{15}{\percent}$, representing a worst-case scenario. 
Tab.~\ref{tab:height} reports the outdoor results, which exhibit the largest absolute errors across our experiments, yet show only minor variation in localization accuracy across all evaluated scenarios.
  \green{During the same flights, an ultrasonic rangefinder on the transmitting \acs{UAV} deviated from the \acs{RTK} height by \SI{4}{\centi\meter} \acs{RMS}, which is below \SI{4}{\percent} of the flight height of the observer and well within this range.}

  To assess sensitivity to surface tilt, we analyzed the localization error for tilt angles $\tau$ up to $\pm$\SI{15}{\degree}.
We physically tilted the observer and bypassed the attitude compensation for the tilt axis, which is equivalent to a level observer over a surface tilted by $\tau$.
We define $\tau<0$ as a tilt that raises the surface toward the transmitter and $\tau>0$ as a tilt away from it.
Fig.~\ref{fig:exp:tilt_surface} shows the theoretical error against range alongside the measured results, which follow the predicted trend.
\green{For negative $\tau$, a singularity occurs at $d_{sing} = h / \tan{|\tau|}$, at which the estimate diverges.
For positive $\tau$, the range is underestimated but remains bounded.
The level-surface assumption thus holds for nearly level surfaces such as calm water, while at $|\tau| = \SI{5}{\degree}$ the error exceeds that of the level-surface experiments beyond about \SI{8}{\meter} (Fig.~\ref{fig:exp:tilt_surface}).}
\section{Discussion}
\label{sec:comparison}
Tab.~\ref{tab:eval:comp} compares our approach against representative state-of-the-art systems.
The \acs{LIDAR}-based approach in~\cite{SwarmLIO2} requires substantially higher onboard compute for point cloud processing than the camera-based methods, and its near-infrared pulses are largely absorbed by water, limiting its applicability over the outdoor water surfaces targeted in this work.
The compared methods also differ in what must be known a priori.
\acs{UVDAR} assumes a known marker geometry, restricting deployment to platforms large enough to carry sufficiently spaced markers. 
CREPES achieves accurate ranging but requires \ac{UWB} transceivers on all team members, adding a radio dependency.
The \acs{CNN}-based approach depends on a training dataset and on consistent lighting conditions.
\green{Unlike onboard stereo cameras, whose baseline is limited by the size of the UAV~\cite{10876012}, the observer and its mirror image form a vertical baseline of $2h$.}
Our method requires only the range to the reflective surface, which is directly measurable onboard, and remains independent of transmitter size and marker configuration, while enabling direct optical identification without reliance on radio communication.

\section{Conclusion}
This work presents an approach for onboard relative localization in \ac{UAV} teams.
Our method estimates the relative position between \acp{UAV} by exploiting typically unwanted reflections of active localization markers attached to team members.
By accounting for uncertainties introduced by camera resolution and surface roughness, the proposed method operates without prior knowledge of team member sizes or optical surface properties.
We validated our approach across indoor and outdoor experiments, demonstrating reliable performance under varying lighting conditions and across different floor materials.
\green{We further characterized the position bias induced by surface tilt.}
Outdoor experiments further showed that our method outperforms \green{the state-of-the-art active-marker system}, particularly at greater distances (\SI{30}{\meter}), reducing the \ac{MAE} by up to \SI{37.3}{\percent} along the optical axis.

\begin{acronym}
        \acro{GNSS}{Global Navigation Satellite System}
        \acro{RTK}[RTK]{Real-Time Kinematic}
        \acro{MAV}[MAV]{Micro Aerial Vehicle}
        \acro{UWB}[UWB]{Ultra-wideband}
        \acro{LED}{Light-emitting diode}
        \acro{UAV}[UAV]{Uncrewed Aerial Vehicle}
        \acro{UGV}[UGV]{Uncrewed Ground Vehicle}
        \acro{USV}[USV]{Uncrewed Surface Vehicle}
        \acro{UVDAR}[\emph{UVDAR}]{UltraViolet Direction And Ranging}
        \acro{CI}[CI]{Confidence Interval}
        \acro{UV}[UV]{Ultraviolet}
        \acro{FOV}{Field of View}
        \acro{LED}{Light-emitting diode}
        \acro{PVC}{Polyvinyl chloride}
        \acro{RF}{radio frequency}
        \acro{OCC}{Optical Camera Communication}
        \acro{LoS}{Line of Sight}
        \acro{OOK}[OOK]{On-off Keying}
        \acro{CNN}[CNN]{Convolutional neural network}
        \acro{MAE}{Mean Absolute Error}
        \acro{LIDAR}[LiDAR]{Light Detection And Ranging}
        \acro{ESS}{Effective Sample Size}
        \acro{RMS}{Root Mean Square}
      \end{acronym}

\bibliographystyle{IEEEtran}
\bibliography{references}

\end{document}